%% file: main.tex
\documentclass[twoside,leqno,twocolumn]{article}
\usepackage{cite}
\usepackage{amsmath,amssymb,amsfonts}
\usepackage{algorithmic}
\usepackage{multirow}
\usepackage{tabu}
\usepackage{url}
\usepackage{tabularx}
\usepackage{makecell}
\usepackage[backref]{hyperref} 
\newcommand{\thickhline}{\noalign{\hrule height 0.8pt}}
\usepackage{graphicx}
\usepackage{textcomp}
\usepackage{xcolor}
\usepackage{caption}

\newcommand*\samethanks[1][\value{footnote}]{\footnotemark[#1]}

\def\BibTeX{{\rm B\kern-.05em{\sc i\kern-.025em b}\kern-.08em
    T\kern-.1667em\lower.7ex\hbox{E}\kern-.125emX}}

\usepackage[letterpaper]{geometry}
\usepackage{ltexpprt}
\usepackage{hyperref}

\begin{document}

\title{Boosting Urban Traffic Speed Prediction via Integrating Implicit Spatial Correlations}

\author{Dongkun Wang\thanks{State Key Laboratory of Internet of Things for Smart City, University of Macau.}
\and Wei Fan\thanks{University of Central Florida.}
\and Pengyang Wang\samethanks[1]
\thanks{Contact Author}
\and Pengfei Wang\thanks{Damo Academy, Alibaba Group}
\and Dongjie Wang\samethanks[2]
\and Denghui Zhang\thanks{Rutgers University}
\and Yanjie Fu\samethanks[2]
}

\date{}

\maketitle


\input{abstract}

\input{introduction}

\input{definition}

\input{methodology}
\input{experiment}
\input{related_work}
\input{conclusion}
\bibliographystyle{plain}   
\bibliography{ref}

\end{document}


\newcommand\relatedversion{}

\title{\Large Appendix\relatedversion}

\date{}

\maketitle

\fancyfoot[R]{\scriptsize{Copyright \textcopyright\ 2023 by SIAM\\
Unauthorized reproduction of this article is prohibited}}

\section{Details about baseline algorithm}
The selected four base models are as follows:

\noindent{\bf (1) TGCN}\cite{zhao2019t}: combining the graph convolutional network (GCN) and gated recurrent unit (GRU) to model spatial and temporal dependency, respectively. 

\noindent{\bf (2) STGCN}\cite{yu2017spatio}: capturing the complex localized spatial-temporal
correlations effectively through an elaborately designed spatial-temporal
synchronous modeling mechanism.

\noindent{\bf (3) STTN}\cite{xu2020spatial}: capturing dynamic spatial dependencies and long-range temporal dependencies for long-term traffic forecasting.

\noindent{\bf (4) DKFN}\cite{chen2020graph}: modeling the self and neighbor dependencies and their predictions are fused under the statistical theory and optimized through the Kalman filtering network.

\section{Implementation}

Out work is implemented based on the framework proposed by \cite{wang2021libcity}. 
Code and raw data are available at 
\protect\url{https://github.com/Dongkun-Wang/SDM2023}.

\bibliographystyle{plain}   
\bibliography{ref}

%% file: abstract.tex
\begin{abstract} \small\baselineskip=9pt
Urban traffic speed prediction aims to estimate the future traffic speed for improving the urban transportation services. 
Enormous efforts have been made on exploiting spatial correlations and temporal dependencies of traffic speed evolving patterns by leveraging explicit spatial relations (geographical proximity) through pre-defined geographical structures ({\it e.g.}, region grids or road networks). 
While achieving promising results, current traffic speed prediction methods still suffer from ignoring implicit spatial correlations (interactions), which cannot be captured by grid/graph convolutions. 
To tackle the challenge, we propose a generic model for enabling the current traffic speed prediction methods to preserve implicit spatial correlations. 
Specifically, we first develop a Dual-Transformer architecture, including a Spatial Transformer and a Temporal Transformer. 
The Spatial Transformer automatically learns the implicit spatial correlations across the road segments beyond the boundary of geographical structures, while the Temporal Transformer aims to capture the dynamic changing patterns of the implicit spatial correlations. 
Then, to further integrate both explicit and implicit spatial correlations, we propose a distillation-style learning framework, in which the existing traffic speed prediction methods are considered as the teacher model, and the proposed Dual-Transformer architectures are considered as the student model. 
The extensive experiments over three real-world datasets indicate significant improvements of our proposed framework over the existing methods.

\end{abstract}

%% file: introduction.tex
\section{Introduction}
\begin{figure}[!tbh]
  \centering
  \vspace{-0.4cm}
  \includegraphics[width=0.9\linewidth]{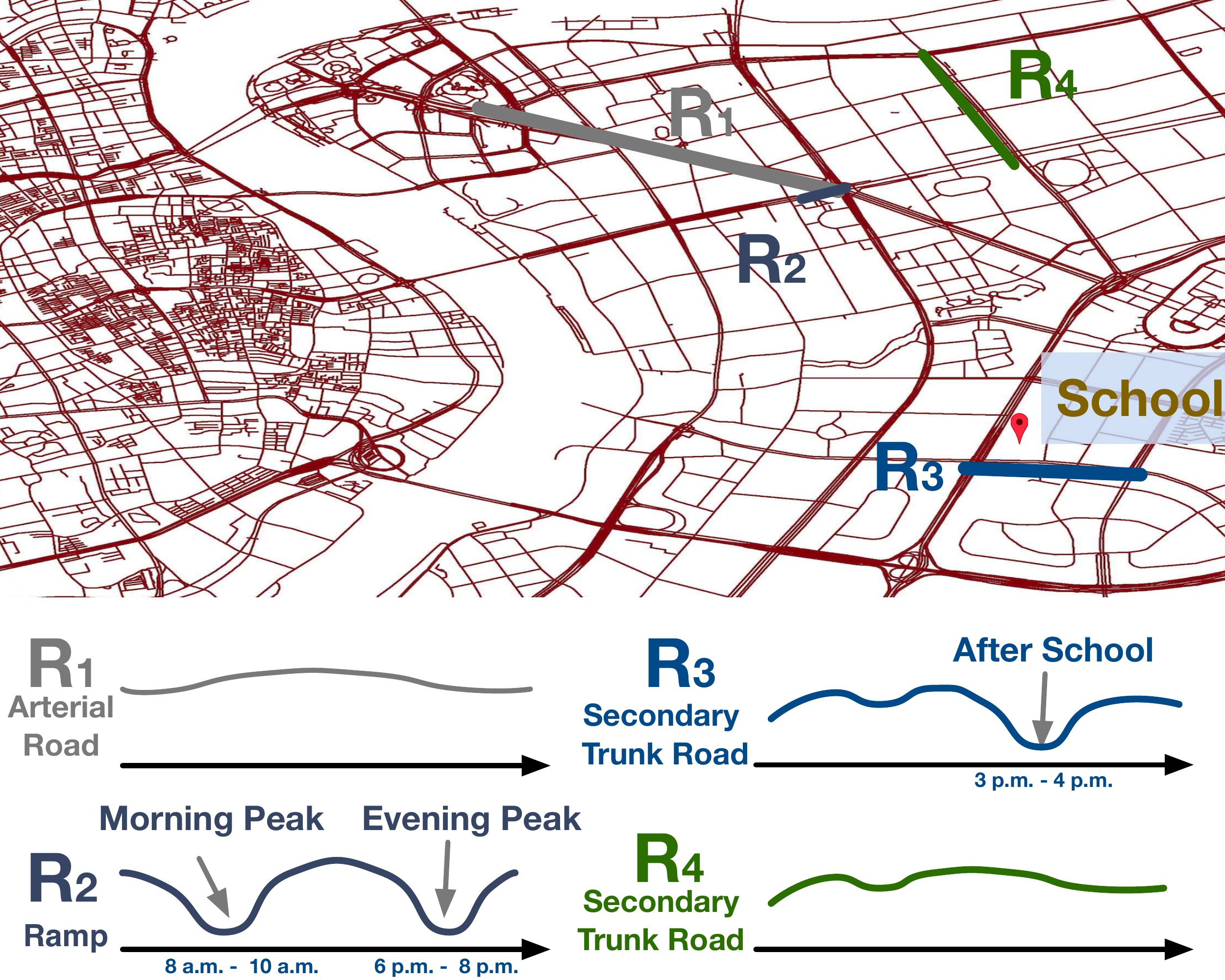}
  \vspace{-0.2cm}
  \caption{An example of implicit spatial correlations for traffic speed in the road network.
  The curves below the road network indicates the traffic speed within 24 hours in a workday. 
  The arterial road $R_1$ and the ramp $R2$ show different speed patterns in morning/evening rush hours, though they are adjacent; the secondary trunk roads $R_3$ and $R_4$ are distant, but sharing similar speed patterns in non-after-school hours. 
  The observations violate the common assumption of spatial correlations: neighboring spots have similar patterns, suggesting implicit spatial correlations existed beyond the geographical proximity.
  }
  \vspace{-0.55cm}
  \label{figure:motivation}
\end{figure}

Facing the explosive growth of urban travel demands with rapid urbanization in recent decades, traffic speed prediction has been playing a vital role in developing intelligent transportation systems (ITSs) in scheduling, planning, and managing urban traffic. 
Traffic speed prediction aims to estimate the future traffic speed based on historical data by exploiting spatial correlations among geographical entities ({\it e.g.}, road segments) and temporal dependencies ({\it e.g.}, temporal impacts of evolving traffic). 
Therefore, how to effectively preserve the spatial correlations and temporal dependencies becomes the crux of achieving accurate traffic speed prediction.

Extensive research has been conducted in studying spatial correlations and temporal dependencies for traffic speed prediction. On the side of temporal dependencies, prior studies either consider the traffic speed prediction as a pure time series ({\it i.e.}, uni/multivariate) forecasting problem~\cite{chandra2009predictions} or exploit deep recurrent neural networks (RNN)~\cite{hochreiter1997long} to capture evolving patterns by updating hidden states~\cite{ma2015long}.
On the side of spatial correlations, the common practice is to investigate relationships between neighbourhood spots. 
Specifically, the whole region is split into small grids, and Convolutional Neural Networks (CNN) are exploited over adjacent regions to examine spatial correlations between surrounding regions~\cite{ke2020two}. 
Beyond the region grids, graph-based models directly take the road network into account, in which road segments are considered as nodes and adjacency relationships between road segments are edges.
Then, Graph Neural Networks (GNNs) are adopted for leveraging the topology of the traffic system via aggregating the traffic patterns of neighbouring connected road segments, which has achieved the state-of-the-art (SOTA) results~\cite{yu2020forecasting} .

Although showing promising performances, existing methods still suffer from the following three unique challenges. 
\textbf{First, implicit spatial correlations are neglected.} 
Spatial correlations are currently modeled through pre-defined geographical structures, such as CNN over the region grids or GNN over road networks, which are in fact explicit spatial correlations that neighboring/adjacent spots should share similar traffic patterns.
However, some contradictory observations arise that neighboring/adjacent spots may indicate distinct patterns. 
Figure~\ref{figure:motivation} shows an example of traffic speed distributions in the city.
During the the morning and evening rush hours, the arterial road $R_1$ maintains faster traffic speed due to larger capacity for the traffic volumes and the regulation for low-speed limit, but the adjacent ramp $R_2$ encounters mild congestion.
On the other hand, surprisingly, even though $R_3$ and $R_4$ are geographically distant, the long-term traffic speed patterns of $R_3$ and $R_4$ are still similar to each other, since $R_3$ and $R_4$ are both the secondary trunk roads and locates in the similar urban functional regions (residential zones).
The phenomenon suggests that there exist implicit spatial correlations beyond the spatial proximity, which are ignored by CNNs or GNNs models. 
Moreover, the over-smoothing issue of GNNs that enforces neighborhoods as similar as possible may even cause nontrivial mistakes. 
Therefore, how to effectively preserve such implicit spatial correlations among road segments remains a challenge.

\textbf{Second, implicit spatial correlations are dynamically changing.} 
Still taking Figure~\ref{figure:motivation} as an example: 
while showing different speed patterns during the morning and evening rush hours, the speed patterns of the arterial road $R_1$ and the adjacent ramp $R_2$ are still closed to each other in the spare time, since the overall traffic is smooth in off-peak hours. 
Moreover, although similar in the long-term, the short-term speed patterns of $R_3$ and $R_4$ are divergent. 
Because one primary school locates at the neighbourhood area of $R_3$, the regulation sets the speed limit as the 20 miles/hour between 3 p.m. and 4 p.m., which leads to the traffic speed of $R_3$ is relative slow even during the off-peak hours. 
Such difference between $R_3$ and $R_4$ would occur periodically (one hour everyday), resulting in short-term divergence of the traffic speed patterns. 
Furthermore, as the intelligent traffic light controllers are developed for optimizing waiting time adaptively, the dynamics of implicit spatial correlations become more complicated. 
Therefore, how to capture the dynamics of implicit spatial correlations is still challenging.

\textbf{Third, the integration of explicit and implicit spatial correlations are not well-studied.} 
As discussed, the explicit spatial correlations have been proven effective in modeling traffic speed patterns, but few efforts are made on studying implicit spatial correlations, much less in the integration. 
In fact, integrating the explicit and implicit spatial correlations is  essential to derive the complete perception of the traffic patterns in terms of the spatial perspective. 
However, the unique characteristics of explicit and implicit spatial correlations hinder the appropriate integration: the explicit spatial correlations are focusing on spatial proximity using pre-defined geographical structures, but the implicit spatial correlations are more complex. 
For the neighboring/adjacent spots, the implicit spatial correlations can still take effect along with the geographical structures but emphasizing on the mutual interaction ({\it e.g.}, $R_1$ interacts with $R_2$ during rush and off-peak hours in Figure~\ref{figure:motivation}); 
for the distant road segments, the implicit spatial correlations break the boundary of the pre-defined geographical structures and enable the interaction between remote but relevant road segments ({\it e.g.}, $R_1$ and $R_3$ are distant but functionally similar Figure~\ref{figure:motivation}). 
Moreover, since the explicit spatial correlations are already well-studied, how to flexibly integrate the implicit spatial correlations with state-of-the-art methods is highly desirable. 

Therefore, to tackle the above challenges, we propose a generic framework for boosting current SOTA traffic speed prediction methods\footnote{Current SOTA methods in traffic speed prediction  are graph-based models that use GNNs to directly learn over road network topology, such as TGCN~\cite{zhao2019t}, STTN~\cite{medrano2021inclusion}, STGCN~\cite{yu2017spatio}, DKFN~\cite{chen2020graph}, etc.} by flexibly integrating implicit spatial correlations. 
Specifically, we first develop a Dual-Transformer architecture to preserve the implicit spatial correlations and the respective dynamic patterns among road segments automatically without utilizing explicit geographic information as prior knowledge . 
To further integrate the explicit and implicit spatial correlations, we devise a knowledge distillation-style learning framework, where we take the SOTA models (explicit spatial correlations captured) as the teacher model and the proposed Dual-Transformer architecture as the student model. 
Along this line, the learned Dual-Transformer architecture will preserve both the explicit and implicit spatial correlations, and thus, boost the performances of the SOTA traffic speed prediction models. 
Our contributions can be summarized as follows:
\begin{itemize}
\itemsep0em 
    \item We introduce the implicit spatial correlations into the urban traffic prediction task to overcome the limitation of the current SOTA methods in over-reliant on the pre-defined geographical structures.
    \item We propose a Dual-Transformer architecture to attack the dynamic changing issue of the implicit spatial correlations across the road segments.
    \item We devise a distillation-style learning framework to flexibly integrate explicit and implicit spatial correlations without bothering to revise the SOTA models.
    \item We conduct extensive experiments on three real-world datasets to validate the effectiveness of our proposed framework.
\end{itemize}

%% file: definition.tex
\begin{figure*}[!t]
	\centering
	\vspace{-0.8cm}
	\includegraphics[width=0.95\linewidth]{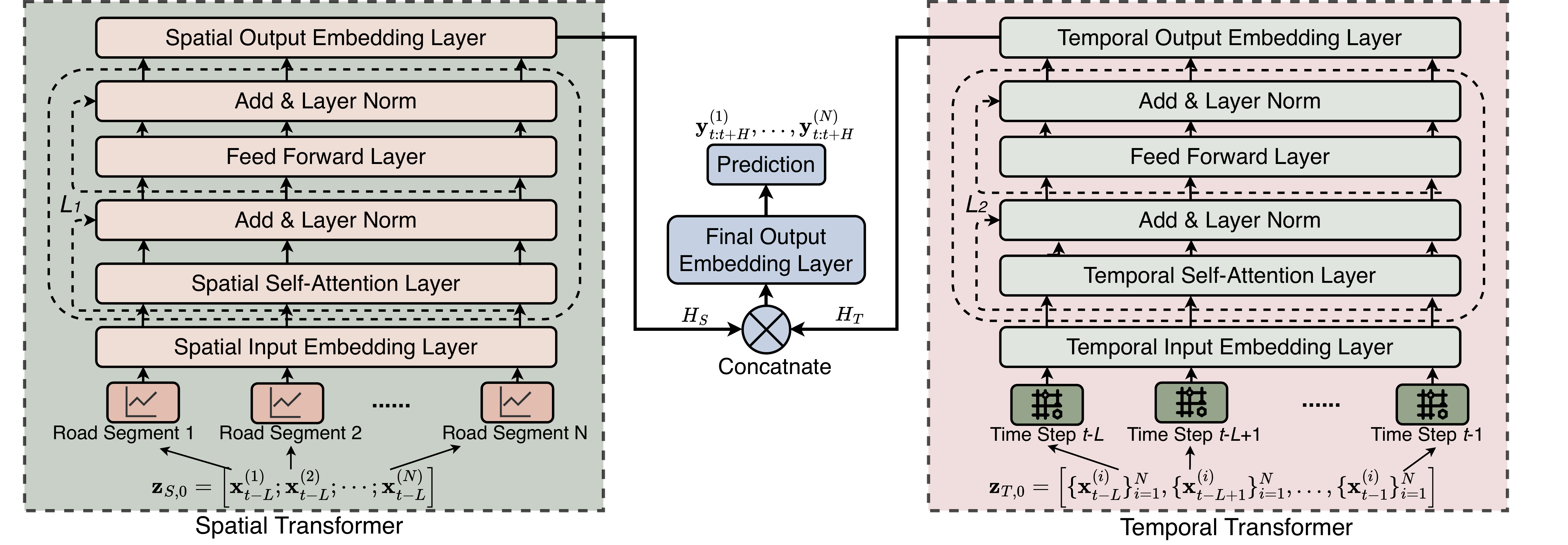}
	 \vspace{-0.5cm}
	 \caption{An illustration of Dual-Transformer.}
	 \vspace{-0.55cm}
	\label{fig:dual}
\end{figure*}

\vspace{-0.5cm}
\section{Problem Formulation}

In this work, we focus on the multi-step traffic speed prediction task with focusing on integrating the explicit and implicit spatial correlations.
Formally, let $\mathcal{G}=(\mathcal{V}, \mathcal{E})$ denote the road network, where $\mathcal{V}=\{v_1, v_2, \dots v_N\}$ represents the road segment set with $N$ segments, and $\mathcal{E}$ is the edge set to demonstrate the adjacency relationship between road segments. 
Each road segment $v_i$ is associated with a $T$-step traffic speed series $\mathbf{x}^{(i)}=\{ x^{(i)}_1, x^{(i)}_2, \cdots, x^{(i)}_T \}$, where ${x}^{(i)}_{t}$ stands for the traffic speed value of $i$-th road segment at $t$-th time step.
Then, the traffic speed records $\mathbf{X}$ of the entire road network $\mathcal{G}$ can be regarded as a multi-variate time series: $\mathcal{X} = \{ \mathbf{x}^{(1)}, \mathbf{x}^{(2)}, \cdots, \mathbf{x}^{(N)}\}$.



Following the classic setting in auto-regressive time series forecasting, given the historical observations $ \mathcal{X}_{t-L:t}=\{ \mathbf{x}^{(1)}_{t-L:t}, \mathbf{x}^{(2)}_{t-L:t}, \cdots,  \mathbf{x}^{(N)}_{t-L:t}  \}$ of a certain time length $L$, we aim to predict future traffic speed in a period of time length $H$, denoted by $\mathcal{X}_{t:t+H}=\{\mathbf{x}^{(1)}_{t:t+H}, \mathbf{x}^{(2)}_{t:t+H}, \cdots, \mathbf{x}^{(N)}_{t:t+H}\}$. Then, the traffic speed prediction problem with the integreation of explicit and implicit spatial correlations can be formulated as 
\begin{equation}
  \mathcal{X}_{t:t+H}  = f(g_1(\mathcal{X}_{t-L:t}, \mathcal{G}), g_2(\mathcal{X}_{t-L:t})) + \epsilon_{t:t+H}
\end{equation}
where $g_1$ is a learnable function to capture the \textbf{explicit spatial correlations} by considering the road network topology, $g_2$ is a learnable function to automatically preserve the \textbf{implicit spatial correlations} without any prior geographical knowledge, $f$ is the integration function, and ${\epsilon}_{t:t+H} = [\epsilon_t, \dots, \epsilon_{t+H-1}]$ denotes a vector of 
{\it i.i.d. }Gaussian noises. 

Noted that since we aim to provide a flexible and generic framework for boosting current SOTA methods, and current SOTA methods have already been working well in capture explicit correlations, we directly adopt the current SOTA methods as $g_1$, and study how to design the implicit spatial correlation function $g_2$ and the integration function $f$. 
Moreover, the current SOTA methods ($g_1$) have inherently captured the temporal dependencies, we will not additionally introduce how to model the temporal dependencies to avoid redundancy.

%% file: methodology.tex
\vspace{-0.2cm}
\section{Methodology}
In this section, we introduce our proposed framework for boosting traffic speed prediction tasks. 
We start with the an overview of the framework. 
Then, we present the details of each component, and summarize the method.

\vspace{-0.3cm}
\subsection{Framework Overview} 
\begin{figure}[!t]
  \centering
  \vspace{-0.2cm}
  \includegraphics[width=1\linewidth]{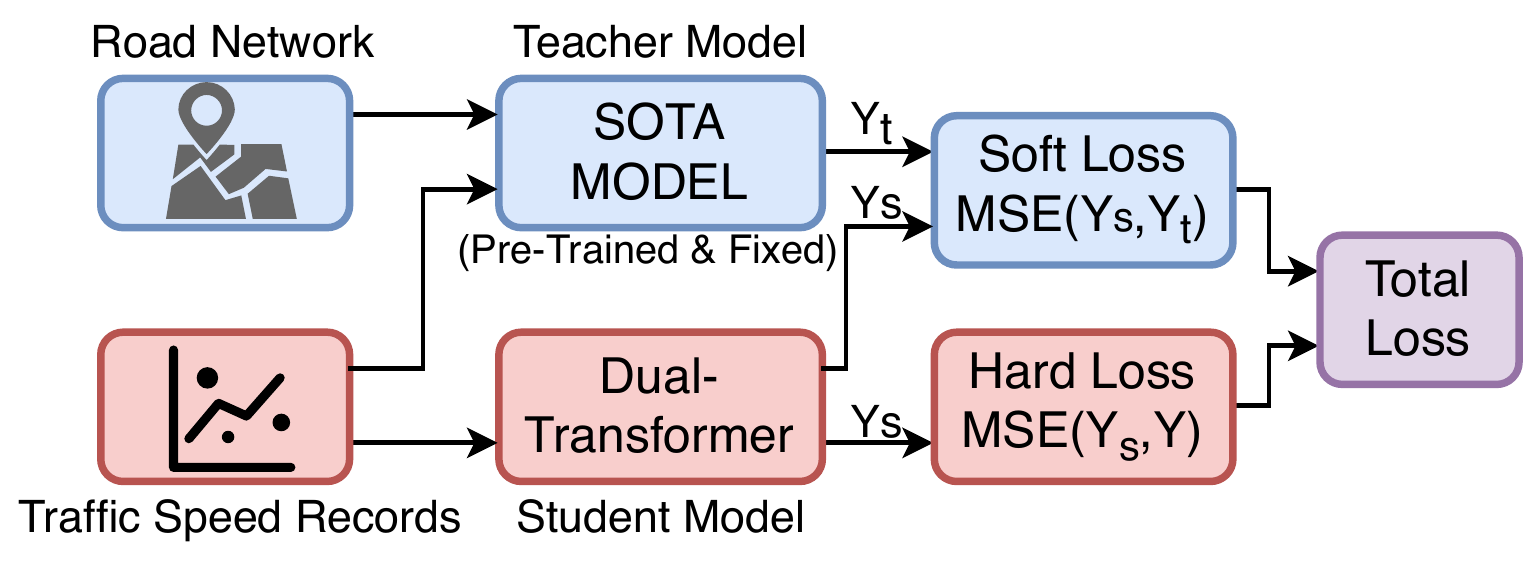}
   \vspace{-0.7cm}
  \caption{An illustration of the knowledge distillation-style integration of explicit and implicit spatial correlations. 
  The SOTA model (pre-trained and fixed) is taken as the teacher model (explicit spatial correlations preserved), while the Dual-Transformer is taken as the student model (implicit spatial correlations preserved).
  The integration is conducted by jointly optimize the ``Soft Loss'' and ``Hard Loss''.}
  \label{figure_KD}
   \vspace{-0.5cm}
\end{figure}

Our proposed framework aims to provide a generic wrapper-style solution to enhance the current SOTA methods by integrating implicit spatial correlations. 
The proposed framework includes two stages: 
(1) preserving implicit spatial correlations, and (2) integrating explicit and implicit spatial correlations. 
Specifically, in Stage I, a Dual-Transformer architecture (as shown in Figure~\ref{fig:dual}) is devised to preserve the implicit spatial correlations and respective dynamics.
In Stage II, a knowledge distillation-style integration framework (as shown in Figure~\ref{figure_KD}) is proposed to integrate explicit and implicit spatial correlations, in which the current SOTA methods are taken as the teacher model, and the proposed Dual-Transformer as the student model. 
The integration framework extracts the knowledge of the explicit spatial correlations from the current SOTA methods, and then pass it into the Dual-Transformer for the integration, finally generating predictions. 
Next, we will introduce the Dual-Transformer and the integration procedure in detail.

\vspace{-0.3cm}
\subsection{Dual-Transformer Architecture for Preserving Implicit Spatial Correlations} 
Transformer architecture exploits the self-attention mechanism to automatically explore the correlations and dependencies among the input tokens~\cite{vaswani2017attention}, showing promising results in nature language processing (NLP) \cite{devlin2018bert} and computer vision (CV)~\cite{dosovitskiy2020image}.
As discussed, the implicit spatial correlations indicate the complex interactions among adjacent road segments and sophisticated similarities among distant entities, which is outside of the traditional geographical proximity modeled by graph-based SOTA methods.
Therefore, Transformer is a promising solution for learning the implicit spatial correlations among road segments. 
Specifically, we design a Dual-Transformer architecture, including a Spatial Transformer for preserving the implicit spatial correlations and a Temporal Transformer for modeling the corresponding dynamics of the correlations.

\noindent \underline{\textbf{Spatial Transformer.}} 
Some efforts have been made to exploit Transformer in spatial-temporal data mining \cite{xu2020spatial, reza2022multi}. 
Regarding time series as an input sequence, an intuitive idea is to treat series data at any time step as a input token to the Transformer.
Specifically, given a lookback window of length $L$, the input sequences are:
\begin{equation}
     \{ \{\mathbf{x}_{t-L}^{(i)}\}_{i=1}^{N}, \{\mathbf{x}_{t-L+1}^{(i)}\}_{i=1}^{N}, ..., \{\mathbf{x}_{t-1}^{(i)}\}_{i=1}^{N} \}
\end{equation}
where $\mathcal{X} \in \mathbb{R}^{L*N}$, which means each time point is considered as a input token and the multi-variate series are considered into the feature dimensions.
This setting is widely adopted by a lot of transformed-based time series forecasting models \cite{zhou2021informer, liu2021pyraformer}, which captures correlations and dependencies of different time steps. However, the important spatial correlations/dependencies are neglected. 
To this end, we propose a Spatial Transformer to conduct the spatial dependency learning and capture the implicit geographic knowledge. We make a very simple yet effective improvement on top of Transformers which achieves strong performances in the traffic speed prediction tasks. 
To directly model correlations among different road segments (also called nodes), we rewrite the aforementioned input of the lookback window by:
\begin{equation}
    \mathcal{X} = \{  \mathbf{x}_{t-L:t}^{(1)}, \mathbf{x}_{t-L:t}^{(2)}, ..., \mathbf{x}_{t-L:t}^{(N)}  \}
\end{equation}
\vspace{-0.2cm} 

\noindent where $\mathbf{x}_{t-L:t}^{(i)} = [ \mathbf{x}_{t-L}^{(i)}, \mathbf{x}_{t-L+1}^{(i)}, ..., \mathbf{x}_{t-1}^{(i)}  ]$ is the traffic speed values from step $t-L$ to step $t-L-1$ of $i$-th road segments;
$\mathcal{X} \in \mathbb{R}^{N*L}$ takes road segments as input tokens and regards the time steps as the feature dimensions. 

Then, we follow the standard Transformer encoder to process and extract implicit spatial correlations, which mainly includes multi-head self-attention (MSA) blocks and multi-perceptron (MLP) blocks. Also, layer-norm (LN) is applied before each block and residual connections are conducted after each block. Specifically, the MSA block can be represented as


\begin{equation}
\begin{aligned}
\operatorname{MSA}(\mathcal{Q},\mathcal{K},\mathcal{V})&= \operatorname{Concat}({head}_{1},\ldots,{head}_{h}) W^{\mathcal{O}} \\
{head}_{i}&= \operatorname{Attention}(\mathcal{Q} W_{i}^{\mathcal{Q}}, \mathcal{K} W_{i}^{\mathcal{K}}, \mathcal{V} W_{i}^{\mathcal{V}}) \\
&=\operatorname{softmax}\left(\frac{\mathcal{Q}W_{i}^{\mathcal{Q}} (\mathcal{K} W_{i}^{\mathcal{K}})^{T}}{\sqrt{d_{k}}}\right) \mathcal{V} W_{i}^{\mathcal{V}}, 
\end{aligned}
\end{equation}

\vspace{-0.2cm} 
\noindent where $\mathcal{Q}$ denotes the query, $\mathcal{K}$ denotes the key, and $\mathcal{V}$ denotes the value, $d_k$ denotes the output dimension; $W_{i}^{\mathcal{Q}}, W_{i}^{\mathcal{K}}, W_{i}^{\mathcal{V}}$ are parameters for query projection, key projection  and value projection; $W_{i}^{\mathcal{O}}$ is the output projection parameters; $h$ denotes the number of heads in multi-head self-attention. 
Here, we set $ \mathcal{Q} = \mathcal{K} = \mathcal{V}  =  \mathcal{X}$ for self-attention.
Then, the Spatial Transformer can be represented as
\begin{equation}
\begin{aligned}
\mathbf{z}_{S,0} &=\left[ \mathbf{x}_{t-L}^{(1)}  ; \mathbf{x}_{t-L}^{(2)}  ; \cdots ; \mathbf{x}_{t-L}^{(N)} \right]\\
\mathbf{z}_{S,\ell}^{\prime} &=\operatorname{MSA}\left(\operatorname{LN}\left(\mathbf{z}_{S,\ell-1}\right)\right)+\mathbf{z}_{S,\ell-1}, \;\;\;\;  \ell=1 \ldots L_1 \\
\mathbf{z}_{S,\ell} &=\operatorname{MLP}\left(\operatorname{LN}\left(\mathbf{z}_{S,\ell}^{\prime}\right)\right)+\mathbf{z}_{S,\ell}^{\prime}, \;\;\;\;  \ell=1 \ldots L_1 \\
\mathbf{H}_{S}  &= \operatorname{LN}\left(\mathbf{z}_{S,L_1} \right)  
\end{aligned}
\end{equation}
where $\mathbf{z}_{\ell}$ is the output after the $\ell$ layers' processing.  The whole encoding process includes total $L_1$ inner layers to finally get the final spatial representations $\mathbf{H}_S$.

\noindent \underline{\textbf{Temporal Transformer.}} 
Although the proposed Spatial Transformer can capture the implicit spatial correlations, another important challenge is that the implicit spatial correlations are dynamically changing over time. 
As aforementioned, the Spatial Transformer regards the road segments (nodes) as the input tokens and captures the implicit spatial relationships. 
However, the Spatial Transformer does not model the temporal correlations among different time steps. In other words, the model cannot sense the dependency changing if the input data has the temporal distribution shift. 

To this end, we develop the Temporal Transformer to preserve the dynamics of implicit spatial correlations.
Specifically, the Temporal Transformer shares the same structure with the Spatial Transformer, but takes the traffic speed data of all the segments in the same time slot as input. 
Formally, the Temporal Transformer can be represented as
\begin{equation}
    \mathbf{z}_{T,0} = \left[ \{\mathbf{x}_{t-L}^{(i)}\}_{i=1}^{N}, \{\mathbf{x}_{t-L+1}^{(i)}\}_{i=1}^{N}, ..., \{\mathbf{x}_{t-1}^{(i)}\}_{i=1}^{N} \right]
\end{equation}
\begin{equation}
    \mathbf{z}_{T,\ell}^{\prime} =\operatorname{MSA}\left(\operatorname{LN}\left(\mathbf{z}_{T,\ell-1}\right)\right)+\mathbf{z}_{T,\ell-1}, \;\;\;\;  \ell=1 \ldots L_2
\end{equation}
\begin{equation}
    \mathbf{z}_{T,\ell} =\operatorname{MLP}\left(\operatorname{LN}\left(\mathbf{z}_{T,\ell}^{\prime}\right)\right)+\mathbf{z}_{T,\ell}^{\prime}, \;\;\;\;  \ell=1 \ldots L_2
\end{equation}
\begin{equation}
    \mathbf{H}_T =\operatorname{LN}\left(\mathbf{z}_{T,L_2} \right)
\end{equation}
where $\mathbf{z}_{T,\ell}$ is the output after the $\ell$ layers' processing.  
The Temporal Transformer also includes total $L_2$ inner layers to get the final temporal representations $\mathbf{H}_T$.

\noindent \underline{\textbf{Combination.}}  
Afterwards, we combine the proposed Spatial and Temporal Transformer to generate the final prediction results.
Specifically, we concatenate the outputs from the Spatial Transformer and the Temporal Transformer, and use an output projection layer to do the prediction, which can be represented as:
\begin{equation}
  \mathbf{y}^{(1)}_{t:t+H}, ..., \mathbf{y}^{(N)}_{t:t+H} = \operatorname{Concat}(\mathbf{H}_{T}, \mathbf{H}_{S})\mathbf{W}_\mathbf{H}
\end{equation}
\vspace{-0.1cm}
where $\mathbf{W}_\mathbf{H}$ is the learnable projection parameters and $\mathbf{y}^{(i)}_{t:t+H}$ are the predicted traffic speed values from step $t$ to $t+H$ at the $i$-th road segments.

The proposed Dual-Transformer architecture pays attention to the implicit spatial correlations and the corresponding dynamics at the same time without using auxiliary geographic information directly. 
Even though the input of the Spatial and Temporal Transformers come from the same origin, these two branches can indicate the pattern behind the data in different feature space. 
This enables the proposed Dual-Transformer more capable to capture implicit spatial dependencies with time-evolving patterns.

\vspace{-0.3cm}
\subsection{Integration of Explicit and Implicit Spatial Correlations}
We propose a novel knowledge distillation-style framework to conduct the integration. 
Specifically, we take the current SOTA methods as the teacher model and the proposed Dual-Transformer as the Student model. 
Intuitively, since the current SOTA methods are graph-based models that utilize GNNs to learn the geographical proximity from the road networks, they can effectively capture the explicit spatial correlations. 
Through knowledge distillation, the explicit spatial correlations are learned and then passed to the Dual-Transformer for the integration. 

Formally, let $Y_t$, $Y_s$, and $Y$ denote the predictions of the SOTA methods (the teacher model), Dual-Transformer (the student model), and the ground-truth of the traffic speed. 
We first pre-train the SOTA methods to fit the ground-truth. 
Then, we fix the SOTA methods, and conduct the integration process by optimizing Dual-Transformer with the SOTA methods. 
Specifically, the integration has two objectives: (1) accepting the knowledge from the SOTA methods, and (2) predicting as accurately as possible. 
Therefore, following the convention of the teacher-student paradigm, the integration process can be represented as: 
\begin{equation}
    Total Loss = \alpha \cdot \underbrace{MSE(Y_{s},Y_{t})}_{\text{Soft Loss}} + \beta \cdot \underbrace{MSE(Y_{s},Y)}_{\text{Hard Loss}} 
    \label{equation:7}
\end{equation}
where $\alpha$ and $\beta$ are fixed weights for Soft Loss and Hard Loss, respectively. 
Specifically, the ``Soft Loss'' is to set the prediction results of the SOTA methods $Y_t$ as the target, and push the prediction of Dual-Transformer $Y_s$ as close as to the SOTA methods. 
Along this line, the learned explicit spatial correlation will be integrated into the Dual-Transformer. 
On the other hand, the ``Hard Loss'' aims to make the Dual-Transformer generate precise prediction results, which can provide the correct optimization direction for the integration. 
The integration is conducted automatically with optimizing the $TotalLoss$ in Equation~(\ref{equation:7}).

\vspace{-0.3cm}
\subsection{Analysis and Discussion}
The key components of the proposed framework lie in two perspectives: (1) Dual-Transformer, which is to preserve the implicit spatial correlations, and (2) knowledge distillation-style learning framework, which is to integrate the explicit and implicit spatial correlations. 
Specifically, the Dual-Transformer has one Spatial Transformer and one Temporal Transformer with the identical architecture. 
The difference is that the Spatial Transformer accepts one road segment with different time-step records, while the Temporal Transformer accepts one time step with all road segments.
The alternation between spatial and temporal dimension is straightforward and simple, but effective in preserving both the implicit spatial correlations and the corresponding dynamics (as validated in Table~\ref{tab:overall} and Figure~\ref{figure:tovis}). 
Then, the integration is realized by knowledge distillation through the teacher-student framework. 
Concatenation or linear combination is unreliable since the explicit and implicit spatial relationships are unknown. 
Therefore, Integrating correlations automatically by optimizing the Dual-Transformer with "Soft Loss" and "Hard Loss" is more promising.
Moreover, the teacher model can be replaced by any graph-based SOTA methods without modifying the inner architecture to fit the framework. 
As Dual-Transformer has fewer parameters and a simpler design than SOTA, it could be better deployed in mobile devices to provide public services.


%% file: experiment.tex
\vspace{-0.3cm}
\section{Experiment}

In this work, we conduct extensive experiments on real-world datasets of Shenzhen, Los Angeles and the San Francisco Bay Area to evaluate the performance of our proposed methods in traffic prediction tasks. Particularly, our experiments aim to answer the following research questions: \\
\noindent \textbf{Q1:} Are implicit spatial correlations indispensable for urban traffic flow prediction? Can our proposed framework boost current SOTA methods? \\
\noindent \textbf{Q2:} How much can explicit and implicit spatial correlations contribute to traffic flow prediction respectively? What percentage to integrate explicit and implicit spatial correlations is optimal? \\
\noindent \textbf{Q3:} How does the Dual-Transformer architecture capture the dynamics of implicit spatial correlations? \\
\noindent \textbf{Q4:} How robust does the Dual-Transformer in preserving the implicit spatial correlations?


To answer the above questions, we first introduce the experiment setting. Then, we present the experimental results and analysis.

\vspace{-0.3cm}
\subsection{Dataset Description}
We evaluate our proposed framework on three real-world datasets~\cite{wang2021libcity} associated with three cities, including: ``SZ\_TAXI'' (Shenzhen), ``METR\_LA'' (Los Angeles), and ``PEMS\_BAY'' (San Francisco).
Table~\ref{table:data_speed} shows the statistics of our traffic speed datasets. 
Each dataset includes Road ID, Adjacency Matrix, Average Traffic Speed, and Time. 
For the proposed Dual-Transformer model, we do not utilize adjacency matrix. In contrast, the teacher models take the adjacency matrix as input and exploit it for explicit spatial correlations.

\begin{table}[t!hbp]
    \vspace{-0.35cm}
	\centering
	\scriptsize
	\tabcolsep 0.01in
	\caption {Statistics of the traffic speed data.}
    \renewcommand\arraystretch{0.7}
	\vspace{-0.3cm}
	\begin{tabular}[t]{c|c|c|c}
		\hline
		\textbf{Name}  & \textbf{ Road Nodes} & \textbf{ Speed Records}  & \textbf{Time Period}  \\ \hline
		SZ\_TAXI & $156$  &$464,256$ & 01/01/2015-31/01/2015 \\ \hline
		METR\_LA & $207$ &$7,094,304$ & 01/03/2012-27/06/2012 \\ \hline
		PEMS\_BAY & $325$ &$16,937,700$ & 01/01/2017-30/06/2017 \\
		\hline
	\end{tabular}
\vspace{-0.35cm}
	\label{table:data_speed}
\end{table}

\begin{table*}[!tbp]
  \centering
  \vspace{-0.8cm}
  \caption{Overall comparison. The best performances are highlighted in bold fonts.}
  \vspace{-0.3cm}
    \begin{tabular}{c|c|c|c|c|c|c|c|c|c}
    \thickhline
    \multirow{2}{*}{Model} & \multicolumn{3}{c|}{SZ\_TAXI} & \multicolumn{3}{c|}{PEMS\_BAY} & \multicolumn{3}{c}{METR\_LA} \\
\cline{2-10}          & MSE   & MAPE  & R2    & MSE   & MAPE  & R2    & MSE   & MAPE  & R2 \\
    \hline
    TGCN  & 18.3972 & 0.0315 & 0.8307 & 23.7210 & 0.0573 & 0.7444 & 162.2813 & 0.1385 & 0.6857 \\
    \hline
    TGCN-DT & \textbf{16.6556} & \textbf{0.0301} & \textbf{0.8467} & \textbf{14.1908} & \textbf{0.0413} & \textbf{0.8471} & \textbf{139.0479} & \textbf{0.1221} & \textbf{0.7307} \\
    \hline
    STTN  & 17.2173 & 0.0247 & 0.8415 & 15.1781 & 0.0404 & 0.8371 & 147.1358 & 0.1146 & 0.7150 \\
    \hline
    STTN-DT & \textbf{16.3054} & \textbf{0.0177} & \textbf{0.8499} & \textbf{14.0113} & \textbf{0.0393} & \textbf{0.8490} & \textbf{137.6050} & \textbf{0.0995} & \textbf{0.7335} \\
    \hline
    STGCN & 17.3385 & 0.0275 & 0.8404 & 16.4577 & 0.0405 & 0.8226 & 148.9741 & 0.1257 & 0.7115 \\
    \hline
    STGCN-DT & \textbf{16.1501} & \textbf{0.0244} & \textbf{0.8514} & \textbf{14.2983} & \textbf{0.0403} & \textbf{0.8459} & \textbf{141.2449} & \textbf{0.1208} & \textbf{0.7264} \\
    \hline
    DKFN  & 23.2648 & 0.0138 & 0.7893 & 24.1551 & 0.0542 & 0.7213 & 204.2863 & 0.1333 & 0.3502 \\
    \hline
    DKFN-DT & \textbf{14.7823} & \textbf{0.0124} & \textbf{0.8619} & \textbf{14.4153} & \textbf{0.0402} & \textbf{0.8333} & \textbf{99.0578} & \textbf{0.1004} & \textbf{0.6849} \\
    \thickhline
    \end{tabular}%
  \label{tab:overall}%
\end{table*}%

\begin{figure*}[!tbh]
  \centering
  \vspace{-0.2cm}
  \includegraphics[width=0.85\linewidth]{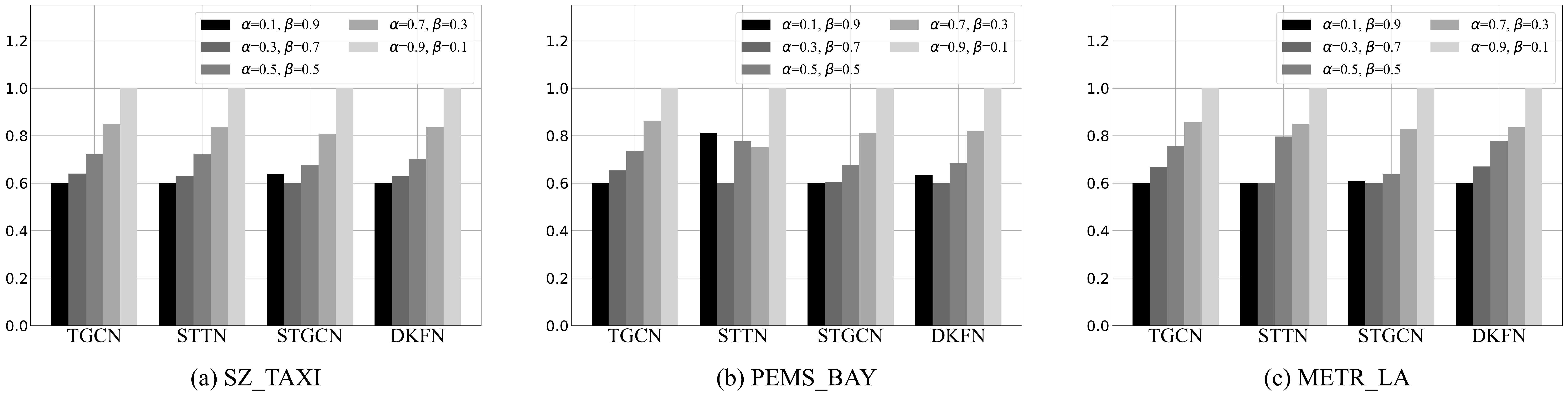}
  \vspace{-0.3cm}
  \caption{An illustration of Dual-Transformer performances {\it w.r.t.} different trade-off parameter pairs. For simplicity, we present the results of normalized MSE in the figure.}
  \label{figure:distill}
  \vspace{-0.35cm}
\end{figure*}

We split the datasets into three non-overlapping sets: for each task, the earliest $70\%$ of the data are the training set, and the following $20\%$ are validating set. The remaining $10\%$ of the data are test set. 



\vspace{-0.3cm}
\subsection{Evaluation Metrics}
Suppose we have $n$ traffic speed prediction tasks to evaluate, let $\hat{y}_{j}$ denote the prediction result, and $y_{j}$ denote the ground-truth for the $j$-th task , respectively. 
We evaluate prediction performances in terms of the following three metrics:

\noindent {\bf (1) $\text{Mean Square Error (MSE)}$:}

$\emph{MSE}=\frac{1}{n} \sum_{j=1}^{n}\left(\hat{y_{j}}-y_{j}\right)^{2}$;

\noindent {\bf (2) $\text{Mean Absolute Percentage Error (MAPE)}$:} 

$\emph{MAPE}=\frac{1}{n} \sum_{i=1}^{n} \frac{\left|\hat{y}_{i}-y_{i}\right|}{y_{i}}$; 

\noindent {\bf (3) $\text{Coefficient of determination
 ($R^{2}$)}$:} 
 
$R^{2}=1-\frac{\sum\left(y_{j}-\hat{y}_{j}\right)^{2}}{\sum\left(y_{j}-\bar{y}\right)^{2}}$, 
where $\bar{y}=\frac{1}{n}\sum \limits_{j=1}^{n} y_j$.

Specifically, for MSE and MAPE, the lower the value, the better the performance; in contrast, for $R^{2}$, the larger the value, the better the performance. 
For all prediction tasks, the time window $T$ for input and time windows $T^{\prime}$ for target are both set to 12.

\vspace{-0.3cm}
\subsection{Comparison Setup}
Since our proposed framework is a generic wrapper for boosting current SOTA  methods with implicit spatial correlations, we evaluate the performance following \textbf{the ablation study manner}. 
Specifically, we take four widely used SOTA methods as the base models, and compare their performance with/without our proposed framework. 
The selected four base models are {\bf (1) TGCN}\cite{zhao2019t}, {\bf (2) STGCN}\cite{yu2017spatio}, {\bf (3) STTN}\cite{xu2020spatial}, and {\bf (4) DKFN}\cite{chen2020graph}.




When applying our proposed framework, we take the base model as the teacher model, and the proposed Dual-Transformer (DT) as the student model. 
We denote the base model powered by our framework as ``*-DT'', where * refers to the base model, such as TGCN-DT, STGCN-DT, STTN-DT, and DKFN-DT, respectively. 
We set the batch size as 128. 
We train the model for 300 epochs with the early-stopping strategy. 
All evaluations were conducted on Ubuntu 18.04.6 LTS, Intel(R) Xeon(R) Gold 6248 CPU, with Tesla-V100 GPU and 128G of memory size.



\vspace{-0.3cm}
\subsection{Q1: Overall Comparison}
We first evaluate the necessity of the implicit spatial correlations and the effectiveness of our proposed the framework. 
We compare the performance of the base model with the corresponding enhanced ``*-DT'' version, where the base model only captures the explicit spatial correlations via preserving spatial proximity, and the ``*-DT'' version additionally integrates implicit spatial correlations. 
The experimental results are shown in Table~\ref{tab:overall}. 
The results indicate that all the enhanced ``*-DT'' versions consistently outperform the base versions across all of the three datasets. 
The results clearly validate our motivation that the implicit spatial correlations are essential for boosting traffic speed prediction. 
Such wrapper-style design benefits the current SOTA models without modifying the original methods, but only need to pass the learned explicit spatial correlations for integration.

\vspace{-0.3cm}
\subsection{Q2: Study of the Contribution of Explicit and Implicit Spatial Correlations}
The contribution of the explicit and implicit spatial correlations is controlled by the effect of the trade-off parameter of Teacher Student Framework, which refer to $\alpha$ and $\beta$ in Equation~\ref{equation:7}.
Specifically, we set $\alpha + \beta = 1$, and select five pair of values, {\it i.e.}, $\{(\alpha=0.1, \beta=0.9), (\alpha=0.3, \beta=0.7), (\alpha=0.5, \beta=0.5), (\alpha=0.7, \beta=0.3), (\alpha=0.9, \beta=0.1)\}$, to investigate the corresponding performances. The larger $\alpha$ is, the more the Dual-Transformer relies on the learned explicit spatial correlations from the SOTA methods. 
We present the normalized MSE for each pair of trade-off parameters in Figure~\ref{figure:distill}.
When $\beta$ is higher than $\alpha$, the model performance is reduced significantly. 
The potential explanation is that although more prior knowledge passed to the Dual-Transformer, over-relying on the SOTA model will distract the Dual-Transformer from producing accurate predictions, resulting in unexpected errors. 
The case will be more obvious when the SOTA methods perform poorly. 
Therefore, a relatively small $\alpha$ (around 0.1-0.3) can guarantee boosting the performance.

\vspace{-0.3cm}
\subsection{Q3: Study of the Dual-Transformer Architecture} 
\begin{figure*}[!tbh]
  \centering
  \vspace{-0.8cm}
  \includegraphics[width=0.85\linewidth]{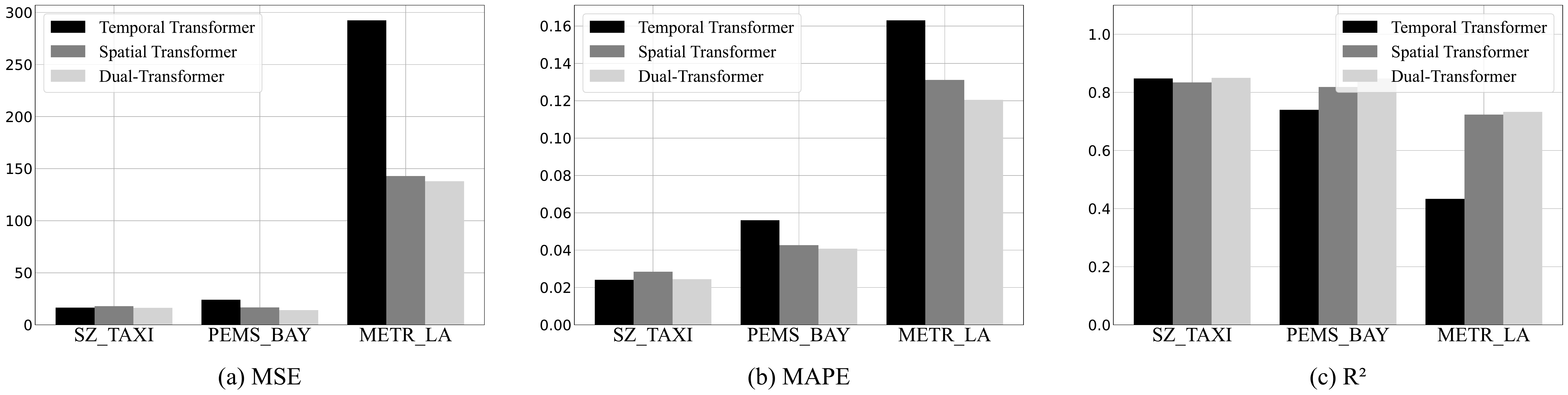}
  \vspace{-0.4cm}
\caption{An ablation study of the proposed Dual-Transformer.}
   \vspace{-0.3cm}
  \label{figure:tovis}
\end{figure*}

In this experiment, we aim to study the necessity of the dynamics of the correlations, and the effectiveness of the design of the Dual-Transformer architecture. 
Specifically, we answer the Q3 in an ablation study manner, where we compare the sole Spatial Transformer (Dual-Transformer without the temporal part), the sole Temporal Transformer (Dual-Transformer without the spatial part), and Dual-Transformer.
Figure~\ref{figure:tovis} shows the results of these three variants of the proposed Dual-Transformer architecture. 
We can observe that among the three variants, the complete version of the Dual-Transformer achieve the best performance. Compared to the Dual-Transformer, the Spatial Transformer only focuses on the implicit spatial correlations, resulting in the deficiency on perceiving the dynamics. 
Such result validates the necessity and effectiveness of preserving the dynamics of the implicit spatial correlations.

\begin{figure*}[!tbh]
  \centering
  \includegraphics[width=0.85\linewidth]{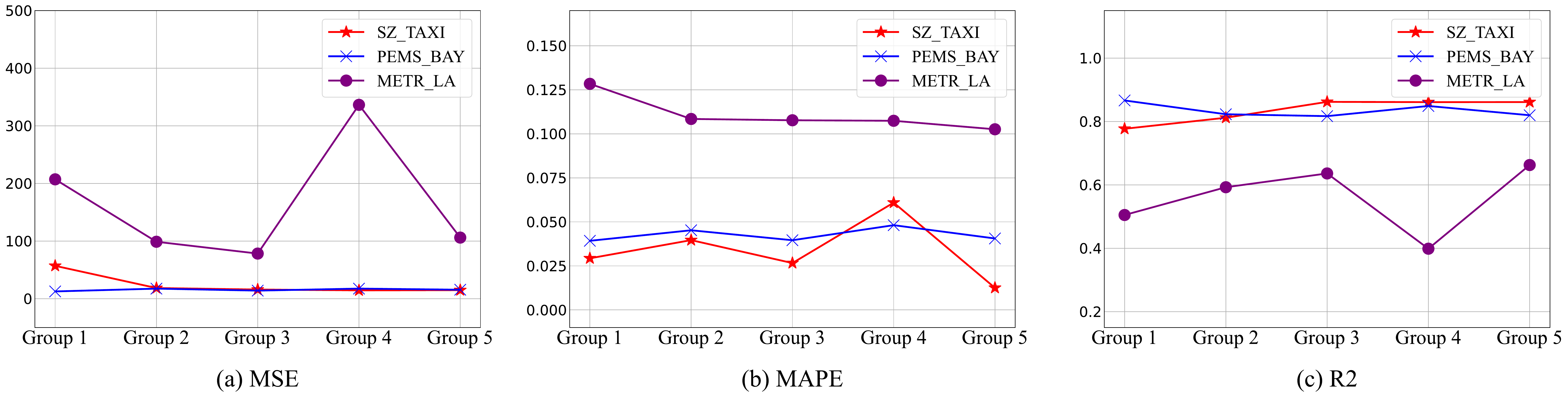}
   \vspace{-0.4cm}
  \caption{An illustration of robustness check of the proposed Dual-Transformer. We evenly split the data into five consecutive and non-overlapped groups. Specifically, for SZ\_TAXI data, the five groups are Jan. 1-Jan. 6, Jan. 7-Jan. 12, Jan. 13- Jan. 18, Jan. 9-Jan.24, and Jan. 25-Jan. 30; for PERMS\_BAY data, the five groups are Mar. 1-Mar. 24, Mar. 25- Apr. 16, Apr. 17-May 11, May 11-Jun. 3, and Jun. 4-Jun. 27; and for MATR\_LA, the five groups are Jan. 1- Feb 5., Feb. 6-Mar. 13, Mar. 14-Apr. 18, Apr. 19-May 24, and May 25-Jun. 30. 
  }
   \vspace{-0.5cm}
  \label{figure:r2_robust}
\end{figure*}

\vspace{-0.3cm}
\subsection{Q4: Robustness Analysis of the Dual-Transformer Architecture}
We also study the robustness of the proposed Dual-Transformer architecture in preserving the implicit spatial correlations. 
We evenly split the data into five non-overlapped time periods.
Then, we evaluated the Dual-Transformer in the five time periods respectively, and examined the variance of the performances. 
We tested the performances in terms of MSE, MAPE and $R^2$. 
The results show that the performance of of the Dual-Transformer is relatively stable on SZ\_TAXI and PEMA\_BAY dataset with small variances. 
An interesting observation standing out is that the performance on METR\_LA dataset is slightly fluctuating. 
A careful inspection of dataset suggests that the data size of METR\_LA is smaller than other two datasets. 
When splitting the dataset into smaller subgroups, the data size is even much smaller, which causes the overfitting issues. 
Comparing the performance on METR\_LA in Figure~\ref{figure:tovis} and Figure~\ref{figure:r2_robust}, the performance on the whole dataset is much better than on each time periods, which corroborates the conjecture of overfitting issue. 
Therefore, the proposed Dual-Transformer is robust in preserving implicit spatial correlations with sufficient data samples, 
but the robustness may be impaired when the input data is limited.

%% file: related_work.tex
\vspace{-0.8cm}
\section{Related work}

\vspace{-0.1cm}
\subsection{Traffic Prediction}
Traffic prediction is initially treated as a time series problem. 
Traditional time series models such as ARIMA \cite{ahmed1979analysis} contribute to forecast traffic data. 
However,they are not capable to model the nonlinear and stochastic features precisely among large-scale traffic data. 
To solve spatial dependence, GCN-based models become popular for traffic prediction. 
Traditional GCN-based models such as T-GCN\cite{zhao2019t} mainly take advantage of the pre-defined geographical structures ({\it e.g.,} region grids or road networks) by maintaining the spatial proximity between adjacent road segments.
Such practice can only capture the explicit spatial correlations. 
Our proposed framework empowers the current SOTA methods with the ability to exploit the implicit spatial correlations by the proposed dual-transformer architecture.

\vspace{-0.3cm}
\subsection{Transformers}
Transformer \cite{vaswani2017attention} has attracted numerous attention of researchers.    
Connecting the encoder and decoder with attention mechanism, Transformer omits the  recurrence and convolutions structure and achieve superior performances in tasks of natural language processing~\cite{devlin2018bert}, where the pre-train \& fine-tuning paradigm becomes the standard pipeline after BERT.
After rapid development in NLP, transformer is also be introduced to computer version domain. 
For example, ~\cite{dosovitskiy2020image} indicates that CNNs is not compulsory to rely on and transformer can also perform well solely on image classification tasks. 
~\cite{touvron2021training} introduced teacher-student framework to transformer and achieve acceptable accuracy. 
Transformer is also widely applied in time series analysis, which utilize the self-attention mechanism or their improvement version with lower time complexity to handle long sequence together. 
Unlike previous applications, in this paper, we introduce Transformer to help preserve the implicit spatial correlations for traffic speed prediction.

\vspace{-0.3cm}
\subsection{Teacher-Student Framework}
Teacher student framework consists of two distinctive parts: teacher model and student model. A well-trained teacher model will transfer information to the student model. There are two common purposes of utilizing teacher student framework:(1) Model Compression. The initial idea is to improve the model compression method proposed by \cite{bucilua2006model}. For classification task, teacher model output represent the probability of classifying the input into different object together with learning the output of intermediate layers, which can make the student to produce a more similar output as the teacher model.
(2) Performance Enhancement. 
Given the prior knowledge from the teacher models, the student models may have better performance than the teacher models. 
\cite{ahn2019variational} propose an creative framework that develop knowledge transfer as maximizing the information betwixt the teacher and the student networks. 

\subsection{Spatial-Temporal Representation Learning}
Spatial-temporal representation learning (STRL) refers to learn quantified features from spatial-temporal data by preserving unique patterns ({\it e.g.,} dynamics, interactions, spatial correlations, temporal dependencies, etc)~\cite{wang2021spatial}, which has been widely deployed in various applications, such as human mobility modeling~\cite{wang2022reinforced,wang2021reinforced,wang2020incremental,wang2019adversarial}, spatial-temporal forecasting~\cite{wang2021automated,wang2021towards,keerthi2020collective,du2019beyond}, urban computing~\cite{wang2021automated,wang2021measuring,wang2020reimagining,wang2020exploiting,zhang2019unifying,fu2019efficient,wang2018learning,fu2018representing,wang2018ensemble}, anomaly detection~\cite{wang2020defending,zhang2016spatial}, transportation~\cite{wang2019spatiotemporal,wang2018you,liu2018modeling,ding2023mst}, agriculture~\cite{liakos2018machine,benos2021machine}, etc. 
Recent works on STRL focus on leveraging the merits of deep learning to discover more complicated relationships among spatial entities. For example, 
\cite{liu2018modeling} proposed a multi-view machine (MVM) method to address the issue of destination prediction in bike-sharing systems (BSSs), including context information from point of interest (POI) data and human mobility data. 
Topic-Enhanced Gaussian Process Aggregation Model (TEGPAM) is a unified probabilistic framework with three components: location disaggregation model, traffic topic model, and traffic speed model. The Gaussian Process model combines new-type data with conventional data~\cite{lin2017road}.
A method was suggested that directly learns the embedding from longitudinal data of users, while concurrently learning a low-dimensional latent space and the temporal development of users in the wellness space~\cite{akbari2017wellness}. \cite{choi2018reinforcement} exploited Markov Decision Process to represent the sequential interactions between users and online recommender systems, and reinforcement learning to impose an optimum policy for providing recommendations.

%% file: conclusion.tex
\vspace{-0.3cm}
\section{Conclusion Remarks}
In this work, we developed a generic wrapper-style framework to boost current SOTA methods by integrating implicit spatial correlations. 
Specifically, we devise a Dual-Transformer architecture with a Spatial Transformer for learning implicit spatial correlations and a Temporal Transformer for capturing the dynamics. 
The explicit spatial correlations are integrated into implicit spatial correlations with a teacher-student learning framework. 
The proposed framework is flexible and can be applied to any current graph-based SOTA methods without any modification. 
The empirical evaluation validates the necessity of implicit spatial correlations and their dynamics, and the effectiveness of our proposed framework for learning such correlations. 

